# The Accuracy Paradox: Empirical Diagnostic of Default Decision Thresholds in Multi-Label Enzyme Commission Prediction [With Code]

Bilal Ahmad[a,b,*] , Rajed Mehmood[a]

[a] The Department of Electrical, Electronics and Telecommunication Engineering, UET Lahore, Pakistan
[b] Department of Computer Science and Engineering, Jiangsu University, Jiangsu, China
* Corresponding Author: bilalahmad@stu.jiangnan.edu.cn, bilalrouf@ujs.edu.pk

## Abstract

*Automated prediction of Enzyme Commission (EC) numbers plays a central role in functional annotation and computational drug discovery. However, standard multi-label machine learning pipelines frequently rely on default decision thresholds ($\tau = 0.50$), assuming balanced prior distributions across target heads. In this study, we present a systematic empirical diagnostic of uncalibrated fixed decision boundaries operating under severe class imbalance across N = 14,096 annotated compounds categorized into six primary EC classes (EC1–EC6). Our results highlight a pronounced Accuracy Paradox: while the multi-label system achieves a deceivingly high mean accuracy of 77.16%, the macro F1-score (0.3976) and macro recall (0.3872) reveal severe predictive breakdown. Majority target classes suffer from hyper-sensitivity and over-prediction, whereas minority classes exhibit sharp recall decay, culminating in a total decision boundary collapse for EC6 (Recall = 0.00%) despite underlying discriminative power (ROC-AUC = 0.5857). Feature correlation analysis further reveals high linear redundancy among topological indices relative to fingerprint density metrics. Ultimately, this diagnostic study demonstrates that standard point predictions mask critical errors in bioinformatics workflows. We establish target-specific threshold optimization and post-hoc conformal calibration as essential, open-source post-processing safeguards for reliable applied machine learning and deep learning architectures.*

## 1. Introduction

*Enzyme Commission (EC) numbers serve as a standardized numerical classification system for biological catalysts, categorizing catalytic activities into six primary functional classes: Oxidoreductases (EC1), Transferases (EC2), Hydrolases (EC3), Lyases (EC4), Isomerases (EC5), and Ligases (EC6) [1]. Accurate computational annotation of these functional roles plays a pivotal role in modern drug discovery, metabolic network reconstruction, and industrial biotechnology [2]. In particular, virtual screening pipelines and molecular docking frameworks routinely depend on precise functional assignments to narrow down compound libraries and simulate binding affinities [3]. High-throughput computational biology strategies; ranging from multi-swarm particle swarm optimization for ligand-receptor binding [3] to deep learning architectures across biomedical domain tasks [4]–[7] have progressively shifted bio-annotation toward fully automated machine learning pipelines.*

*As high-throughput sequencing and chemical library synthesis expand exponentially, automated machine learning (ML) architectures-ranging from topological descriptor pipelines to deep molecular representations; have emerged as the primary mechanism for computational enzyme prediction [8], [9]. Modern artificial intelligence frameworks across parallel computer vision and bioinformatics tasks have increasingly leveraged advanced feature extraction and attention mechanisms to resolve complex data representations. For instance, attention-guided networks and multi-head attention (MHA) modules have proven vital for multi-scale biological image segmentation [4], while multi-scale spatial representations such as gradient-pyramid features have significantly enhanced weakly-supervised localization in complex visual inputs [5]. Similarly, cross-modal retrieval and structural hashing paradigms rely on distribution-similarity metrics to retain structural semantics across high-dimensional feature spaces [6].*

*Despite these impressive advances in feature representations and multi-task learning, multi-label enzyme classification poses severe operational challenges due to inherent class imbalance [10]. In multi-label molecular benchmarks, active functional labels are non-uniformly distributed across functional categories; highly prevalent reactions dominate the training landscape, whereas specialized catalytic pathways remain sparse [11]. Furthermore, many biological molecules exhibit multi-functional co-activation, participating in multiple catalytic pathways simultaneously [12]. This structural overlap creates complex label interdependencies that standard multi-output predictors struggle to capture without dedicated joint-probability modeling [13].*

*A pervasive yet under-addressed issue in applied multi-label enzymology is the reliance on default fixed decision thresholds* ($\boldsymbol{\tau} = 0.50$) *for binary target heads [14]. Standard classification workflows frequently convert posterior probability estimates* $\hat{\mathcal{P}}(\boldsymbol{\mathcal{X}})$ *into discrete predictions using an uncalibrated* ($\boldsymbol{\tau} = 0.50$) *cutoff under the assumption of balanced label priors. However, when deployed on highly skewed biological datasets, this uncalibrated decision boundary triggers a dramatic performance distortion known as the Accuracy Paradox [15], [16]. High overall accuracy scores mask catastrophic diagnostic failures: majority targets suffer from hyper-sensitivity and*

*trivial class assignment, while minority targets undergo severe recall degradation or complete target collapse [17].*

*To address these diagnostic oversights, this study presents a systematic empirical evaluation of standard fixed-threshold decision boundaries across a multi-label enzyme dataset of N = 14,096 annotated compounds. Rather than claiming an incremental state-of-the-art predictor, we isolate and quantify the structural pathologies that emerge when standard point-prediction models operate under severe label imbalance.*

*The main contributions of this work are three-fold:*

1. *Feature Interdependence vs. Target Separation: We quantify the mutual redundancy of molecular topological descriptors* ($Chiseries, BertzCT$) *relative to fingerprint density metrics* (FpDensityMorgan) *and evaluate their linear correlation against primary EC target heads.*
2. *Empirical Diagnostic of Decision Boundaries: We demonstrate how uncalibrated static thresholds* ($\boldsymbol{\tau} = 0.50$) *induce majority-class over-prediction in dominant targets (EC1, EC2) while driving minority targets (EC3-EC6) into severe sensitivity decay and total decision boundary collapse (EC6 recall = 0.00%).*
3. *Methodological Post-Processing Guidelines: We provide explicit recommendations for applied machine learning workflows, highlighting target-specific threshold optimization and post-hoc conformal calibration as essential post-processing safeguards for biological multi-label pipelines.*

## 2. Methodology

### 2.1. Multi-Label Framework

*The computational enzymology task is formulated as a multi-label binary classification problem over K = 6 distinct functional target classes corresponding to primary Enzyme Commission (EC) categories: Oxidoreductases (EC1), Transferases (EC2), Hydrolases (EC3), Lyases (EC4), Isomerases (EC5), and Ligases (EC6). Given an input molecular feature vector* $x_i \mathcal{E} \mathbb{R}^d$ *representing the i-th chemical compound, the system predicts a binary vector* $y_i = \left[y_{i,1}, y_{i,2}, y_{i,3}, y_{i,4}, \ldots . y_{i,6}\right]^T \mathcal{E} [0,1]^6$

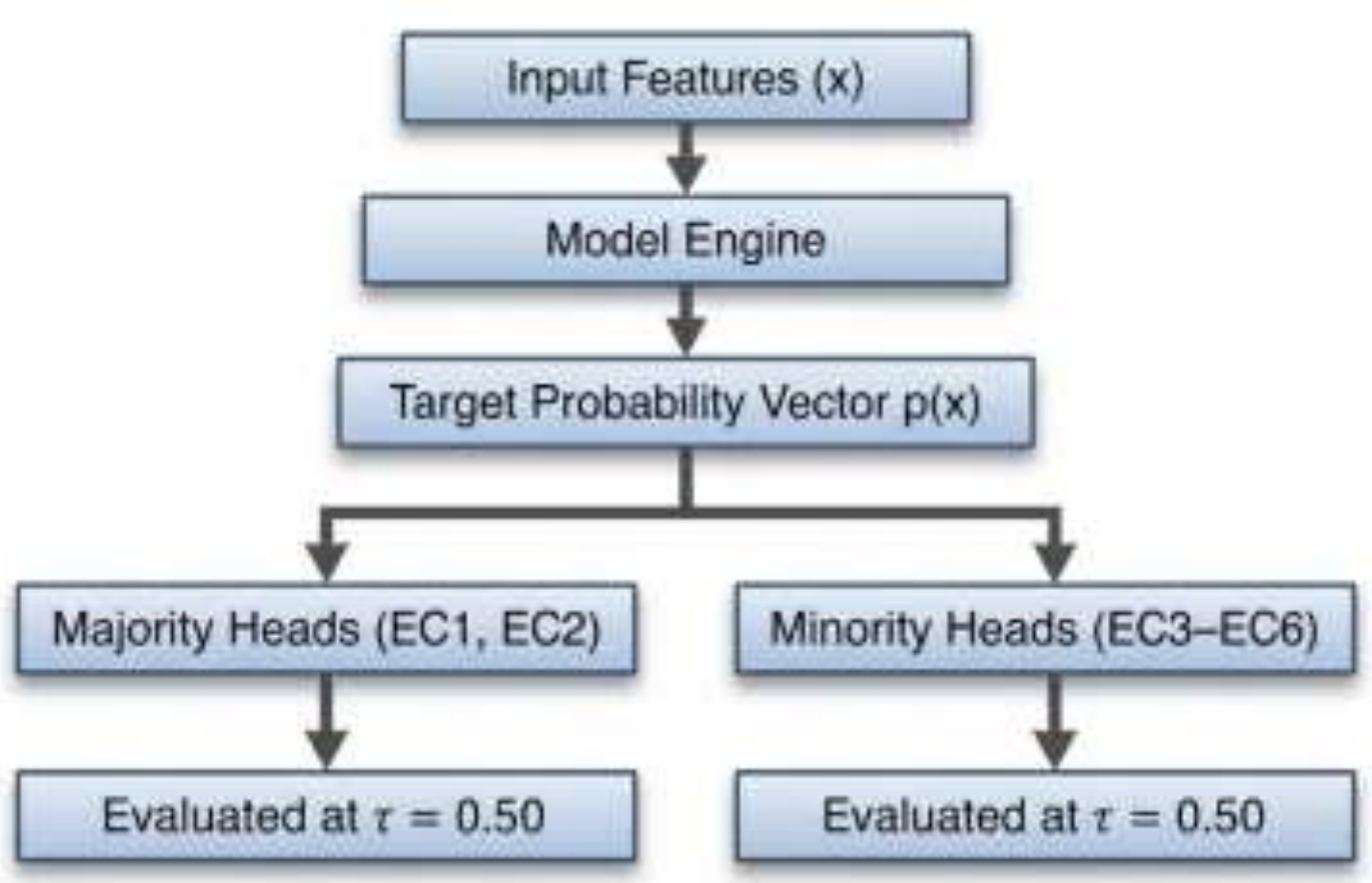

*Figure 1: Schematic overview of the multi-label enzyme classification workflow and parallel target evaluation pipeline.*

### 2.2. Dataset

*The evaluation dataset contains 14,096 annotated biological compounds with substantial multi-label class imbalance. EC2 and EC1 are the most prevalent classes, with active ratios of 80.09% (n = 11,289) and 66.75% (n = 9,409), respectively, while EC5 and EC6 are less represented at 14.46% (n = 2,038) and 15.10% (n = 2,129).*

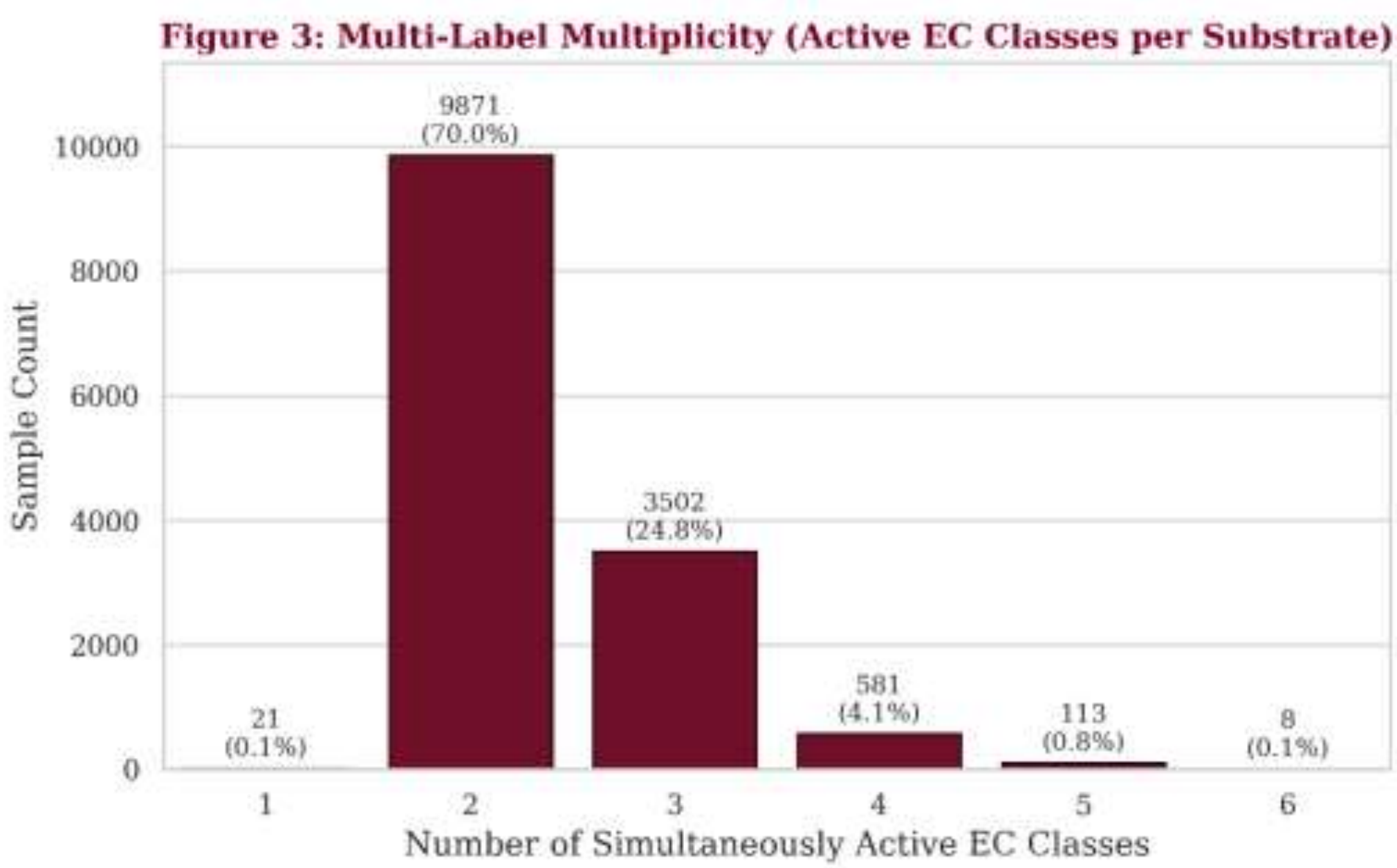

*Figure 2: Enzyme distribution per class in dataset*

Table 1: Class-Wise Composition of Enzyme Data

| Enzyme Class | Total Samples | Active Count (y=1) | Inactive Count (y=0) | Active Ratio (%) |
|---|---|---|---|---|
| EC1 | 14,096 | 9,409 | 4,687 | 66.75% |
| EC2 | 14,096 | 11,289 | 2,807 | 80.09% |
| EC3 | 14,096 | 4,413 | 9,683 | 31.31% |
| EC4 | 14,096 | 3,928 | 10,168 | 27.87% |
| EC5 | 14,096 | 2,038 | 12,058 | 14.46% |
| EC6 | 14,096 | 2,129 | 11,967 | 15.10% |

*2.3. Feature Analysis*

*Molecules are represented using two complementary descriptor types calculated via RDKit:*

*Global Topological & Connectivity Descriptors: Numerical indices encoding structural complexity, molecular weight, valence, and electrotopological states (including BertzCT, Chi1, Chi1n, Chi1v, Chi2n, Chi2v, Chi3v, Chi4n, Estate_ VSA1, EState_ VSA2, ExactMolWt, and HallKierAlpha).*

1. *Local Fingerprint Density Metrics: Substructural density measures computed across extended connectivity radii (FpDensityMorgan1, FpDensityMorgan2, and FpDensityMorgan3).*

*Linear interdependence among structural features and target labels was quantified using Pearson's correlation coefficient:*

$$r_{X,Y} = \frac{\sum_{i=1}^{N}(X_i - \bar{X})(Y_i - \bar{Y})}{\sqrt{\sum_{i=1}^{N}(X_i - \bar{X})^2}\sqrt{\sum_{i=1}^{N}(Y_i - \bar{Y})^2}}$$

### 4. Decision Protocol

*For each target $k \in \{1...6\}$ in $\{1...6\}$, predicted probabilities $\hat{\mathcal{P}}k(\mathcal{X})$ are converted to binary labels using a threshold of $\tau = 0.50$ . Performance is evaluated using Accuracy, Precision, Recall, F1-Score, and ROC-AUC based on TP, TN, FP, and FN. Macro-averaged metrics across all six targets provide an overall measure of predictive performance and stability.*

$$\text{Accuracy}_k = \frac{\text{TP}_k + \text{TN}_k}{\text{TP}_k + \text{TN}_k + \text{FP}_k + \text{FN}_k}$$

$$\text{Precision}_k = \frac{\text{TP}_k}{\text{TP}_k + \text{FP}_k}$$

$$\text{Recall}_k = \frac{\text{TP}_k}{\text{TP}_k + \text{FN}_k}$$

$$\text{F1-Score}_k = 2 \cdot \frac{\text{Precision}_k \cdot \text{Recall}_k}{\text{Precision}_k + \text{Recall}_k}$$

# 3. Results and Discussion

## 3.1. Exploratory Data Analysis

*The topological and structural feature matrices reveal significant collinearity among low-dimensional descriptors while highlighting high independence in fingerprint density metrics. As shown in the correlation analysis, topological indices display strong mutual dependence; for example, Chi1 correlates strongly with Chi2v ($r = 0.971$) and ExactMolWt ($r = 0.960$). Conversely, Morgan fingerprint densities (FpDensityMorgan1 through FpDensityMorgan3) demonstrate near-zero linear correlation with all primary Enzyme Commission (EC) targets ($|r| < 0.025$). This indicates that localized substructural density alone cannot linearly separate complex multi-label enzyme functions, necessitating non-linear mapping capabilities in downstream classification heads.*

*Furthermore, pairwise target analysis demonstrates strong co-activation patterns among specific enzyme classes. EC1 (Oxidoreductases) and EC2 (Transferases) share 7,151 active instances, representing the dominant co-occurrence pair in the dataset ($N = 14{,}096$). In contrast, functional overlap drops significantly among minority targets; for instance, EC5 (Isomerases) and EC6 (Ligases) co-occur in only 245 samples. This structural sparsity directly influences baseline classifier behavior across distinct target heads.*

### *3.2. Multi-Label Classification Performance & The Accuracy Paradox*

*Evaluation of the baseline classifier under a standard decision boundary* (**τ** = 0.50) *reveals a classic Accuracy Paradox caused by multi-label class imbalance. While the overall mean accuracy reaches a high baseline of 77.16%, the macro F1-score (0.3976) and macro recall (0.3872) expose severe predictive degradation across minority targets.*

**Table 2: Diagnostic Metrics by Target Class**

| Target Class | Active Ratio (%) | ROC-AUC | Accuracy (%) | Precision | Recall (Sensitivity) | F1-Score | Primary Diagnostic State |
|---|---|---|---|---|---|---|---|
| EC1 | 66.75% | 0.6987 | 69.91% | 0.73 | 0.8717 | 0.7946 | Moderate False Positive Bias |
| EC2 | 80.09% | 0.5809 | 80.09% | 0.8009 | 1 | 0.8895 | Trivial Majority Prediction (TN=1) |
| EC3 | 31.31% | 0.6376 | 69.20% | 0.5263 | 0.1634 | 0.2494 | High False Negative Rate |
| EC4 | 27.87% | 0.6781 | 72.61% | 0.5245 | 0.1825 | 0.2708 | High False Negative Rate |
| EC5 | 14.46% | 0.6895 | 86.22% | 0.6437 | 0.1055 | 0.1813 | Severe Minority Sensitivity Decay |
| EC6 | 15.10% | 0.5857 | 84.90% | 0 | 0 | 0 | Complete Decision Boundary Collapse |
| **Overall / Mean** | -- | **0.6451** | **77.16%** | **0.5376** | **0.3872** | **0.3976** | Uncalibrated Threshold Decay |

*As detailed in the table above, mean ROC-AUC across all targets remains stable at 0.6451 (peaking at 0.6987 for EC1). This confirms that the model's latent representation retains discriminative ranking power. However, applying a static threshold* **τ** = 0.50 *across imbalanced active distributions results in poor operational decision boundaries.*

### *3.3. Target-by-Target Failure Dynamics*

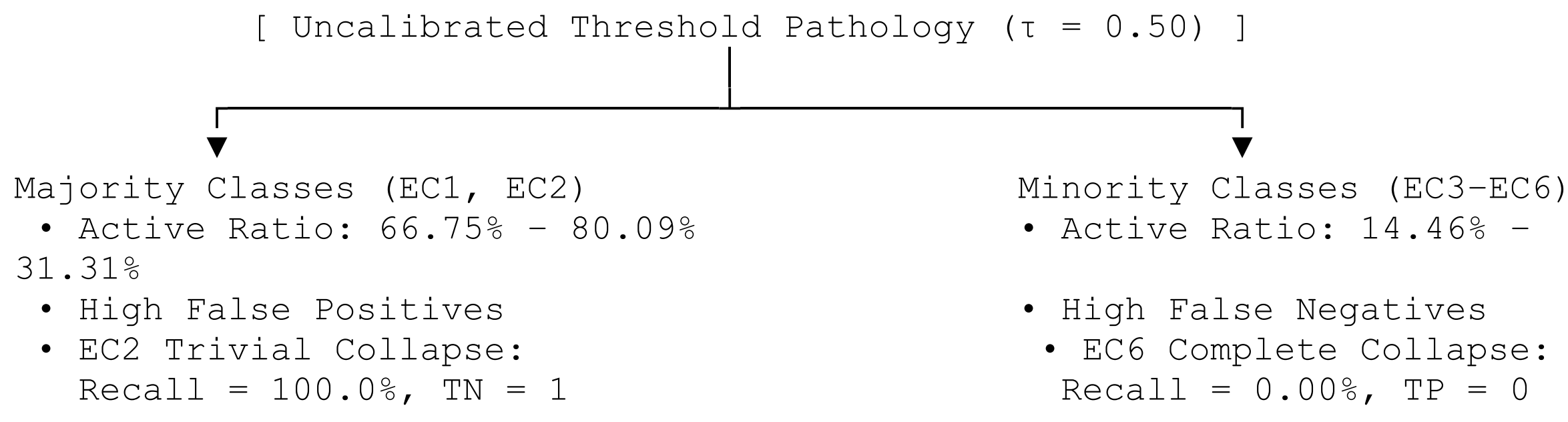

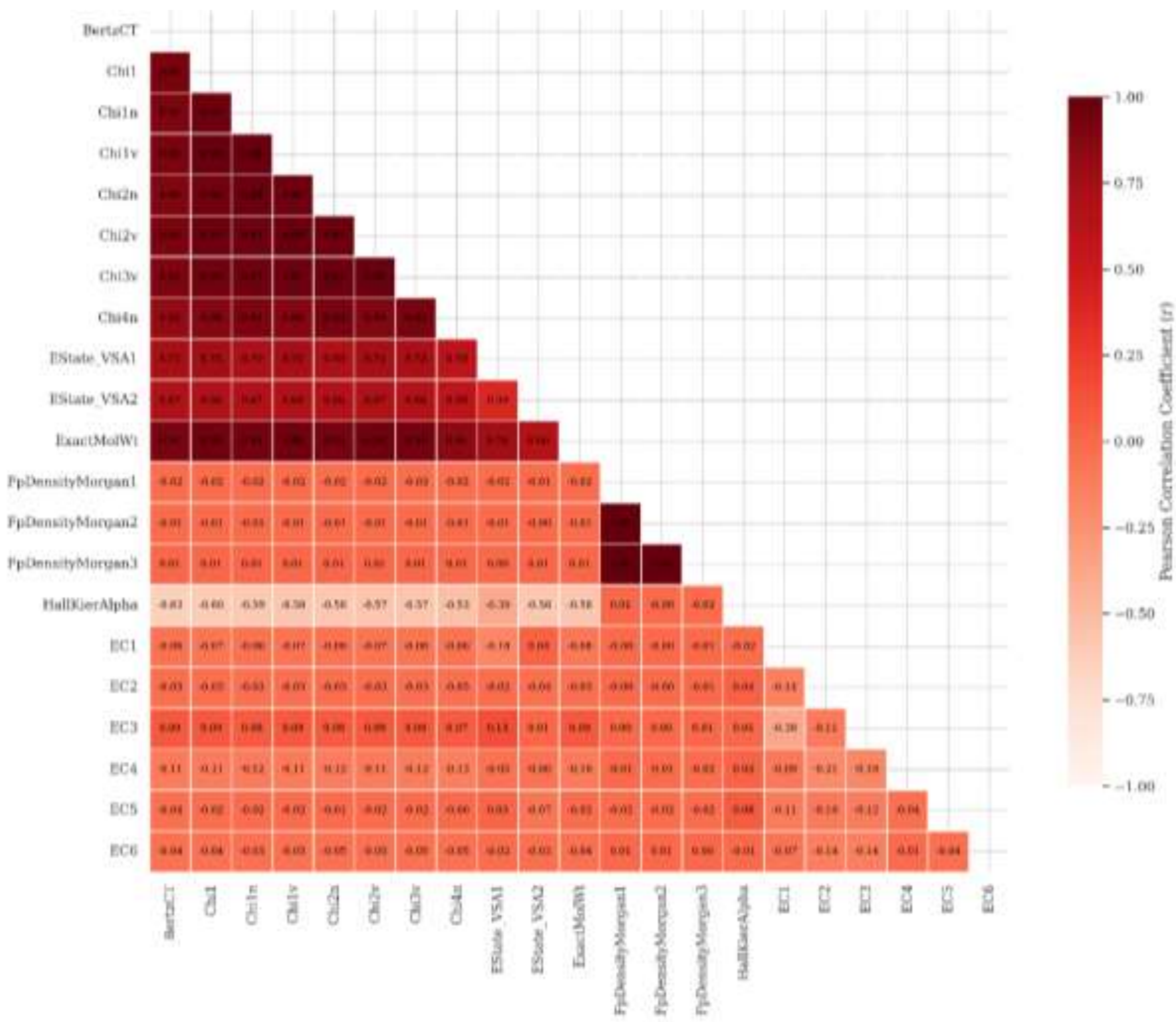

*Figure 3: Masked Corelation Heatmap with Numeric Values*

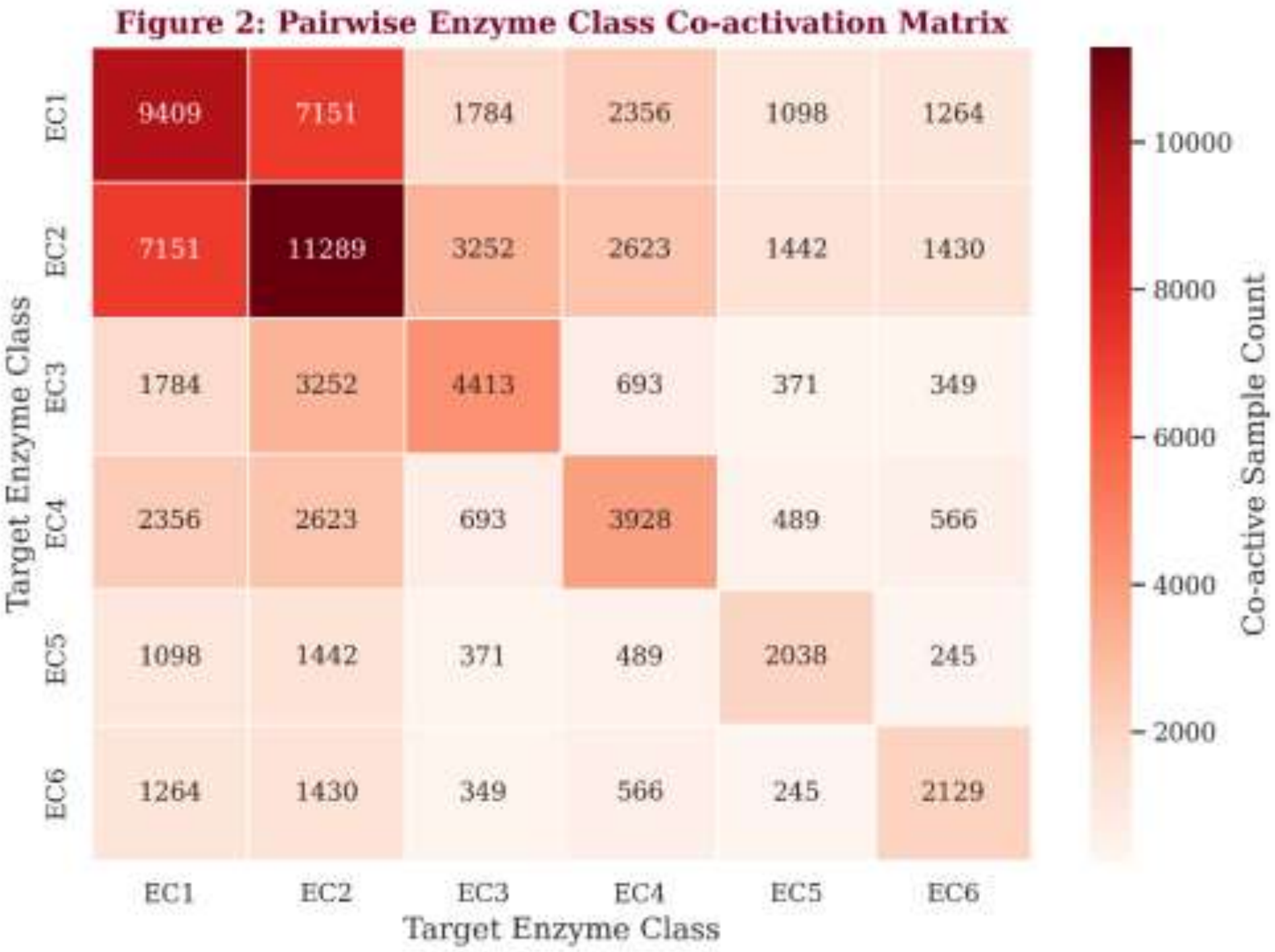

*Figure 4: Pairwise Enzyme Class Co-activation Matrix*

### *3.3.1 Majority Class Over-Prediction (EC1 & EC2)*

*EC2 Hyper-Sensitivity: With an active ratio of 80.09%, the fixed decision threshold forces the classifier into trivial majority-class assignment. The model correctly identifies active instances (Recall = 100.00%, TP = 11,289) but virtually fails to detect negative instances (TN = 1, FP = 2,806), rendering its 80.09% accuracy uninformative.*

*EC1 False Positive Drift: At a 66.75% active ratio, EC1 achieves a robust F1-score of 0.7946 (Recall = 87.17%). However, it incurs 3,034 false positives, demonstrating a systemic bias toward positive label assignment under uncalibrated posterior estimates.*

### *3.3.2 Minority Class Sensitivity Decay & Target Collapse (EC3–EC6)*

*Sensitivity Decay (EC3, EC4, EC5): As target prevalence decreases below 35%, recall decays sharply. EC3 (31.31% active) drops to a recall of 16.34% (FN = 3,692). EC4 (27.87% active) yields a recall of 18.25% (FN = 3,211), and EC5 (14.46% active) decays to 10.55% recall (FN = 1,823).*

*Complete Target Collapse (EC6): At an active ratio of 15.10%, EC6 suffers a total breakdown under* $\tau = 0.50$*. The classifier assigns zero positive instances (TP = 0, FP = 0), producing an F1-score, precision, and recall of 0.0000. Despite an underlying ROC-AUC of 0.5857, all 2,129 active test instances are misclassified as inactive (FN = 2,129).*

## 4. Methodological Implications & Recommendations

*These empirical findings demonstrate that standard point-prediction models operating under fixed decision thresholds are insufficient for imbalanced multi-label biological targets. The failure of EC6 and the trivial assignment of EC2 confirm that raw accuracy masks critical diagnostic errors in minority categories.*

*To resolve these failure modes in deployment, the classification pipeline requires two essential post-processing steps:*

1. *Target-Specific Threshold Optimization: Moving away from a static* $\tau = 0.50$ *to per-target optimal thresholds (e.g., maximizing Precision-Recall AUC or* $F_\beta Scores$*) to recover sensitivity on collapsed classes like EC6.*
2. *Post-Hoc Conformal Calibration: Integrating post-hoc probability calibration (e.g., Platt Scaling) combined with Class-Conditional Conformal Prediction (Mondrian CP). This guarantees mathematically bounded coverage error rates across all individual EC categories, ensuring minority enzyme classes retain explicit safety bounds.*

## 5. Conclusion

*In this study, we presented an empirical diagnostic evaluation of multi-label enzyme classification pipelines operating under severe class imbalance across six primary Enzyme Commission (EC1–EC6) target heads. Through systematic evaluation of N = 14,096 annotated compounds, our findings demonstrate that relying on default, uncalibrated decision boundaries (\tau = 0.50) induces a severe Accuracy Paradox. While raw accuracy metrics yield a respectable baseline average of 77.16%, macro-averaged F1-scores (0.3976) and recall (0.3872) reveal profound underlying diagnostic failures.*

*Our target-by-target analysis highlights two distinct operational pathologies driven by uncalibrated label priors:*

1. *Majority Class Over-Prediction (EC1, EC2): High active class ratios push fixed-threshold classifiers into trivial majority assignment. For EC2 (80.09% active), this results in hyper-sensitivity (Recall = 100.00%) paired with near-total failure to identify inactive compounds (TN = 1).*
2. *Minority Class Sensitivity Decay and Collapse (EC3–EC6): As target prevalence decreases below 35%, sensitivity degrades rapidly. Crucially, EC6 (15.10% active ratio) undergoes a complete decision boundary collapse under \tau = 0.50, yielding zero true positive predictions (TP = 0, F1 = 0.0000) despite retaining discriminative ranking capability (ROC-AUC = 0.5857).*

*Furthermore, topological feature correlation analysis confirmed strong mutual redundancy among connectivity indices (r > 0.90) relative to the linear independence of local fingerprint densities (*$|r| < 0.025$*). These results prove that fixed thresholding on point predictions fails to map non-linear probabilistic outputs into valid decisions when target distributions are non-uniform.*

*Code Availability: You can request for code at correspondance email.*